\documentclass[sigconf,authorversion,nonacm]{acmart}
\usepackage{graphicx}
\usepackage{booktabs}
\usepackage{multirow}
\usepackage{subfig}
\usepackage{hyperref}
\usepackage{pifont}
\usepackage{amsmath}
\usepackage{xcolor}
\usepackage{setspace}
\usepackage[ruled,linesnumbered]{algorithm2e}

\newcommand*{\MyIndent}{\hspace*{0.7cm}}
\newcommand{\cmark}{\ding{51}}%
\newcommand{\xmark}{\ding{55}}%

\AtBeginDocument{%
  \providecommand\BibTeX{{%
    \normalfont B\kern-0.5em{\scshape i\kern-0.25em b}\kern-0.8em\TeX}}}

\setcopyright{acmcopyright}
\copyrightyear{2021}
\acmYear{2021}
\acmDOI{10.1145/XXXXXX.XXXXX}

\acmConference[CIKM '22]{CIKM '22: 31st ACM International Conference on Information and Knowledge Management}{October 17-22, 2022}{Atlanta, Georgia, USA}
\acmBooktitle{CIKM '22: 31st ACM International Conference on Information and Knowledge Management, October 17-22, 2022, Hybrid Conference, Hosted in Atlanta, Georgia, USA}
\acmPrice{15.00}
\acmISBN{978-1-4503-XXXX-X/18/06}

\begin{document}

\title{An Answer Verbalization Dataset for Conversational Question Answerings over Knowledge Graphs}


\author{Endri Kacupaj}
\email{kacupaj@cs.uni-bonn.de}
\orcid{0000-0001-5012-0420}
\affiliation{%
  \institution{University of Bonn, Germany}
  \country{}
}

\author{Kuldeep Singh}
\email{kuldeep.singh1@cerence.com}
\affiliation{%
  \institution{Zerotha Research and Cerence GmbH, Germany}
  \country{}
}

\author{Maria Maleshkova}
\email{maria.maleshkova@uni-siegen.de}
\affiliation{%
  \institution{University of Siegen, Germany}
  \country{}
}

\author{Jens Lehmann}
\email{jens.lehmann@cs.uni-bonn.de}
\affiliation{%
 \institution{University of Bonn, Germany}
 \country{}
}

\renewcommand{\shortauthors}{Kacupaj, et al.}

\begin{abstract}
    We introduce a new dataset for conversational question answering over Knowledge Graphs (KGs) with verbalized answers. Question answering over KGs is currently focused on answer generation for single-turn questions (KGQA) or multiple-tun conversational question answering (ConvQA). However, in a real-world scenario (e.g., voice assistants such as Siri, Alexa, and Google Assistant), users prefer verbalized answers. This paper contributes to the state-of-the-art by extending an existing ConvQA dataset with multiple paraphrased verbalized answers. We perform experiments with five sequence-to-sequence models on generating answer responses while maintaining grammatical correctness.
    We additionally perform an error analysis that details the rates of models' mispredictions in specified categories. Our proposed dataset (Verbal-ConvQuestions) extended with answer verbalization is publicly \textbf{available} with detailed documentation on its usage for wider \textbf{utility}\footnote{\url{https://github.com/endrikacupaj/Verbal-ConvQuestions}}.
\end{abstract}



\maketitle

\section{Introduction}
Question answering (QA) over Knowledge Graphs (KGs) is an essential task that maps a user's utterance to a query over a KG in order to retrieve the correct answer~\cite{singh2018reinvent}. Recently, with the increasing popularity of intelligent personal assistants, the research focus of the scientific community has shifted to conversational question answering over KGs (ConvQA) within multi-turn dialogues~\cite{christmann2019look,kacupaj2021lasagne,plepi2021context}.

\begin{figure}[!t]
\centering
\captionsetup{type=figure}
\includegraphics[width=0.47\textwidth]{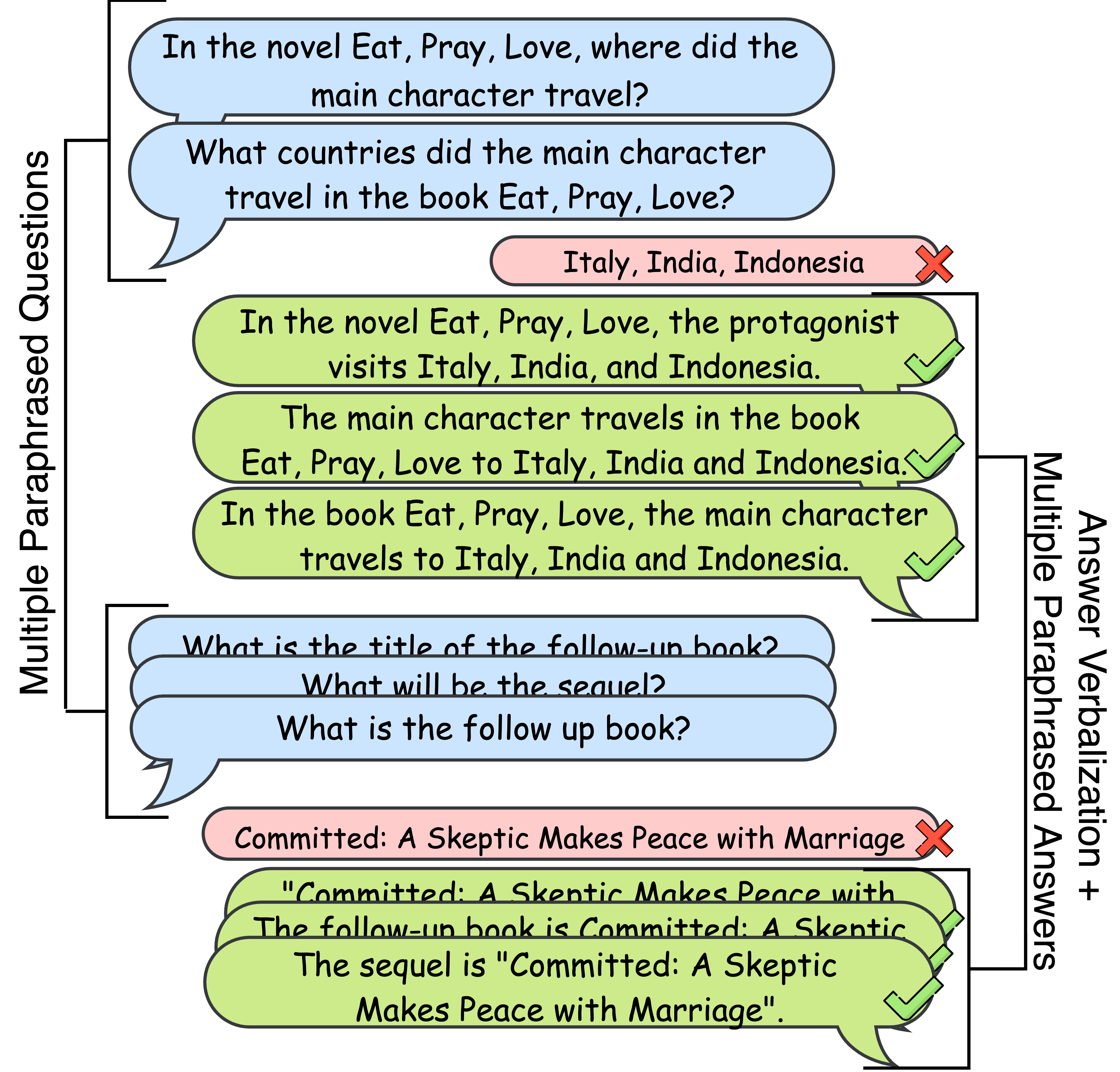}
\caption{An example of multiple verbalized answers and paraphrased questions as contributions of our work (in green and in sky-blue). We use questions from ConvQuestions~\cite{christmann2019look} that provide non-verbalized answers (in red).}
\label{fig:chat_examples}
\end{figure}

\begin{table*}[!ht]
\centering
\resizebox{\textwidth}{!}{%
\begin{tabular}{lccccc}
\toprule
\textbf{Dataset} & \textbf{Large-scale (>=10k)} & \textbf{Complex Questions} & \textbf{Paraphrased Questions} & \textbf{Verbalized Answers} & \textbf{Paraphrased Answers} \\ \midrule
CSQA~\cite{saha2018csqa} & \cmark & \cmark & \xmark & \xmark & \xmark \\
ConvQuestions~\cite{christmann2019look} & \cmark & \cmark & \xmark & \xmark & \xmark \\
ConvRef~\cite{conquer2021kaiser} & \cmark & \cmark & \cmark & \xmark & \xmark \\
\midrule
Verbal-ConvQuestions (\textbf{Ours}) & \cmark & \cmark & \cmark & \cmark & \cmark \\
\bottomrule
\end{tabular}%
}
\caption{Comparison of Verbal-ConvQuestions with existing conversational KGQA datasets. The lack of answer verbalization and paraphrased utterances remains a key gap in the literature, which is a key novelty of proposed dataset.}
\label{tab:comparison}
\end{table*}

However, existing open-source KGQA systems are restricted to only generating or producing answers without verbalizing them in natural language~\cite{fu2020survey}. The lack of verbalization makes the interaction with the user unnatural and often leaves the users with ambiguity~\cite{einolghozati2020sound}. Let us consider the first question in Figure~\ref{fig:chat_examples}, ``What countries did the main character travel in the book Eat, Pray, Love?", where existing QA systems will only respond with the countries as an answer, with no further explanation. In such cases, the user might need to refer to external data sources to verify the answer. Nevertheless, a verbalized answer would allow the user to confirm that the answer is related to the context, since it also includes additional characteristics that indicate how it was determined. For example, in task oriented dialogues, verbalized answers are a common phenomenon~\cite{peskov2019multi}. 

Recent efforts introduce answer verbalization in a QA dataset~\cite{kacupaj2021paraqa,biswas2021vanilla}. The associated empirical results indicate the effectiveness of answer verbalization~\cite{kacupaj2021paraqa,kacupaj2021vogue}. Albeit effective,
answer verbalization for ConvQA is an open research direction due to unavailability of dataset(s) (cf. Table \ref{tab:comparison}). This is precisely one of the reasons why this critical information retrieval task is narrowly studied \cite{conquer2021kaiser}.

Hence, in this paper, we address the task of answer verbalization in conversational question answering over knowledge graphs. In particular, we extend an existing English ConvQA dataset \cite{christmann2019look} with multiple paraphrased natural language answers, using a semi-automated framework that employs back-translation as a paraphrasing technique.
Moreover, we perform experiments with various models on generating the verbalized answer.
We further provide an analysis to detail the reasons for erroneous cases. Overall, we make the following key contributions as \textbf{novelty}:

\begin{itemize}
    \item We provide the first ConvQA dataset over knowledge graphs with multiple verbalized answers. We also supplement the dataset with up to three paraphrased questions. 
    \item We empirically illustrate the quality of the generated verbalization by employing several sequence-to-sequence models.
\end{itemize}














Following is the structure of the paper: Section~\ref{sec:approach} provides details of the conversational dataset and the process for extending it with multiple verbalized answers. Section~\ref{sec:experiments} describes the experiments, including the modeling approaches, the experimental setup, the results, and the error analysis. Section~\ref{sec:related_work} summarizes the related work. Finally, we conclude in Section~\ref{sec:conclusions}.
\section{Data-Generation and Augmentation}\label{sec:approach}
We inherit questions from ConvQuestions~\cite{christmann2019look}, which is a high quality and large-scale benchmark for conversational QA over Wikidata KG~\cite{vrande2014wikidata}. The dataset contains 11,200 conversations, and it was compiled from inputs of crowd workers on Amazon Mechanical Turk with conversations from five domains: Music, Soccer, TV Series, Books, and Movies. The questions incorporate challenging phenomenons such as aggregations, compositionality, temporal reasoning, and comparisons. 

\begin{table}[t]
\centering
\begin{tabular}{l|ccc}
\toprule
\textbf{Dataset} & \textbf{Train} & \textbf{Val.} & \textbf{Test} \\ \midrule
\# Conversations & 6,720 & 2,240 & 2,240 \\
\# Paraphrased Question & 68,447 & 22,368 & 22,400 \\
\# Paraphrased Answer & 68,447 & 22,368 & 22,400 \\
Avg. Question length & 8.48 & 8.75 & 8.01 \\
Avg. Answer length & 8.82 & 9.19 & 8.39 \\ \bottomrule
\end{tabular}
\caption{Verbal-ConvQuestions statistics.}
\label{tab:stats}
\end{table}

\subsection{Initial Answer Verbalization}
The first step is to generate the initial answer verbalization from the seed answers given in the original dataset. We group all similar questions or question templates and reword them using a rule-based approach. To maintain consistency across all answers, we cover the question and answer entities using a few general tokens ($ENT, ANS$). We substitute the tokens back to the original position after the first version is generated. Similarly to other works~\cite{einolghozati2020sound}, we use box brackets to distinguish the seed answer from the remaining sentence; this is helpful when experimenting with the verbalized answers. For example, for the question ``What countries did the main character travel in the book Eat, Pray, Love?" the step would provide an initial verbalized answer: ``The main character travels in the book Eat, Pray, Love to [Italy, India, and Indonesia].'' Here, the answer is mainly a paraphrased version of the question that includes the seed answers.

\begin{table*}[ht]
\centering
\def\arraystretch{1.1}
\begin{tabular}{l|cccccccccc}
\toprule
\textbf{Domain} & \multicolumn{2}{c}{\textbf{Books}} & \multicolumn{2}{c}{\textbf{Music}} & \multicolumn{2}{c}{\textbf{Movies}} & \multicolumn{2}{c}{\textbf{TV Series}} & \multicolumn{2}{c}{\textbf{Soccer}} \\ \midrule
\textbf{Models} & \multicolumn{1}{l}{\textbf{BLEU-4}} & \multicolumn{1}{l}{\textbf{METEOR}} & \multicolumn{1}{l}{\textbf{BLEU-4}} & \multicolumn{1}{l}{\textbf{METEOR}} & \multicolumn{1}{l}{\textbf{BLEU-4}} & \multicolumn{1}{l}{\textbf{METEOR}} & \multicolumn{1}{l}{\textbf{BLEU-4}} & \multicolumn{1}{l}{\textbf{METEOR}} & \multicolumn{1}{l}{\textbf{BLEU-4}} & \multicolumn{1}{l}{\textbf{METEOR}} \\ \midrule
CNNSeq2seq & 0.1331 & 0.3303 & 0.2388 & 0.5199 & 0.1176 & 0.4534 & 0.1151 & 0.4595 & 0.0556 & 0.4040 \\ 
Transformer & 0.3751 & 0.6893 & 0.3594 & 0.6892 & 0.2633 & 0.6496 & 0.2301 & 0.6310 & 0.1995 & 0.5801 \\ 
BERTSeq2seq & 0.4231 & 0.7266 & \underline{0.4403} & \underline{0.7620} & 0.3117 & 0.6815 & 0.3063 & 0.6799 & 0.2396 & 0.6150 \\ 
BART & \textbf{0.6088} & \textbf{0.8577} & \textbf{0.6524} & \textbf{0.8962} & \textbf{0.5925} & \textbf{0.8268} & \textbf{0.5416} & \textbf{0.8484} & \textbf{0.4616} & \textbf{0.7132} \\ 
T5 & \underline{0.5540} & \underline{0.8264} & 0.4033 & 0.7435 & \underline{0.3974} & \underline{0.7542} & \underline{0.4799} & \underline{0.8156} & \underline{0.3323} & \underline{0.6626} \\
\bottomrule
\end{tabular}
\caption{Baselines on Verbal-ConvQuestions dataset. We report BLEU-4 and METEOR scores for generating the verbalized answer. The best result is in bold and the second best is underlined.}
\label{tab:results}
\end{table*}

\subsection{Paraphrasing Questions \& Answers}
Once we generate the first verbalized answer, inspired by \cite{federmann2019multilingual,edunov2018understanding}, we employ back-translation to produce multiple instances for the paraphrased answer. In particular, the back-translation is done using a transformer-based model \cite{vaswani2017attention} as translators. We produce paraphrases using translators for two sets of languages (English-Russian, English-German). For the selection, we considered the models' performance on the WMT'18 dataset \cite{bojar2018proceedings}, including translations between different languages.
The same paraphrasing process was applied to the rest of the conversational questions of the dataset. As output, we get paraphrased verbalized answers and questions.

\subsection{Result Validation}
Finally, to ensure the grammatical correctness of all the generated questions and answers, we include two rounds of a peer-review process. Similar to \cite{einolghozati2020sound}, the first round included a set of annotators where we asked them to assert the produced results and rephrase them if needed. This step ensures more natural and fluent dialogues. Next, another set of annotators were asked to validate the previous step and rephrase if needed. In particular, for the initial verbalization, 55\% required some human interventions. While for paraphrased utterances, 47\% required corrections during the peer-reviewed process.
Table~\ref{tab:stats} indicates statistics of our generated dataset for all three sets. 

\section{Experiments}\label{sec:experiments}
\subsection{Modeling Approaches}
While the proposed dataset refers to the new task of "KGQA with verbalized answers", our experiments focus only on generating the verbalized answers using the following inputs 1) the current question utterance and 2) conversation history. We assume the seed answer is given. Hence, we intend to study the ability of different models in generating the verbalized answer instead of producing the correct seed answer. We employ the following models:

\begin{itemize}
    \item CNNSeq2seq~\cite{gehring2017conv}: a convolutional based encoder-decoder model.
    \item Transformers~\cite{vaswani2017attention}: a model based solely on attention mechanism.
    \item BERTSeq2seq~\cite{rothe2020bertseq2seq}: a BERT \cite{devlin2019bert} encoder-decoder model.
    \item BART~\cite{lewis2020bart}: a de-noising auto encoder that uses a bidirectional encoder and a left-to-right decoder.
    \item T5~\cite{colin2019t5}: an encoder-decoder model pretrained on a mixture of unsupervised and supervised tasks.
\end{itemize}

Based on our dataset, further research in the community is expected to focus on developing end-to-end systems covering both the QA (generating seed answer) and answer verbalization task.
\subsection{Experimental Setup}
\textbf{Prepossessing}
For enabling the models to access the information from past conversational turns, we have to incorporate the dialogue history as an additional input to the models. For the same, we collect all previous utterances (questions and verbalized answers). Together with the current question, we concatenate them using a unique $SEP$ token to allow the model to differentiate between the utterances and information from any previous turn. 

\noindent\textbf{Configurations}
For training, we employ a batch size of $16$, a learning rate of $5e-5$, and we train for $10$ epochs where we store the models' checkpoint with the lowest loss. For the optimization, we use the AdamW, which implements the Adam \cite{kingma2015adam} algorithm with weight decay fix as introduced in \cite{loshchilov2017decoupled}. During optimization, we clip the gradients with a max norm of $1$. We further apply a dropout with a probability $0.1$ across all the models. For criterion, we use cross entropy loss, which combines log softmax and negative log-likelihood loss.
For CNNSeq2seq and Transformer, we employed an embedding dimension of $d=512$ and $l=2$ layers, using standalone implementations.
We use the Transformers Python library~\cite{wolf2020transformers} for experimenting with the remaining models (BERTSeq2seq, BART, T5), and we apply hyperparameters and configurations as introduced in the original papers.

\noindent\textbf{Evaluation Metrics}
We use standard metrics for machine translation and NLG tasks: the BLEU score~\cite{papineni2002bleu} examines the co-occurrences of n-grams in the reference and the proposed responses. It computes the n-gram precision for the whole dataset, which is then multiplied by a brevity penalty to penalize short translations. We report results for BLEU-4 in our experiments. The METEOR score~\cite{banerjee2005meteor} refers to the harmonic mean of uni-gram precision and recall, with recall weighted higher than precision. Metrics can be in the range of $0.0$ and $1.0$.

\subsection{Results}
Table~\ref{tab:results} summarizes the results of the baseline models. We perceive that BART significantly outperforms the other baselines on all five domains for both the BLEU-4 and METEOR scores. The T5 model performs equally well. The main reason is that both models, BART and T5, have been pretrained for several generation tasks. 
Interestingly, all models perform worse in the ``Soccer'' domain; this occurs due to similar terms (e.g. ``plays for") it shares with other domains (e.g. ``Movies" and ``Books"). Furthermore, all models perform relatively well on domains such as ``Music" and ``Books."
As stated, the presented results only refer to generating verbalized answers using the conversational context as an input to the models. The results indicate promising performance and support the creation of the dataset.

\begin{table}[!h]
\centering
\begin{tabular}{l|cccccccccc}
\toprule
\textbf{Models} & \textbf{Grammatical} & \textbf{Semantic} & \textbf{Entity Related} \\ \midrule
CNN & 86\% & 91\% & 77\%  \\ 
Transf. & 41\% & 48\% & 35\% \\ 
BERT & 33\% & 41\% & 32\%  \\ 
BART & 16\% & 19\% & 14\% \\ 
T5 & 18\% & 23\% & 21\% \\
\bottomrule
\end{tabular}
\caption{Error rate of baseline models.}
\label{tab:error}
\vspace{-2em}
\end{table}

\subsection{Error Analysis}
We randomly sampled 200 incorrect model predictions. We detail the error rates for three different categories. The first one, ``Grammatical", refers to examples where the verbalized answer can be understood but contain some grammatical errors such as a mismatch between the noun and verb forms.
Next, the ``Semantic" error class refers to the cases where the primary meaning of the answer has changed; this can occur by introducing new content (e.g., entities) or by omitting essential parts of the content. ``Entity Related" refers to generated answers where the model failed to copy the correct entities from the input utterance or replaces them with pronouns. Table \ref{tab:error} presents the error rates for the three categories. We can observe that the BART and T5 models contain the lowest error rates, and therefore, validate their superior performance. Our empirical study provides a glance at the different types of errors that may occur while targeting the task of KGQA with verbalized answers.

\subsection{Case Study}
We manually examine some examples from the proposed dataset and discuss the intentions behind their construction. 
Regularly, all the required information is provided with the question/query for the question answering task. Therefore, to generate verbalized answers in such scenarios, we have to concentrate only on the question context, considering that the seed answer is given. However, in conversational question answering, we have scenarios such as anaphora and ellipsis \cite{saha2018csqa,christmann2019look}, where the context from previous turns has to be incorporated in order to answer the given question. 
Hence, we had to consider all these cases when creating the dataset. 
Table~\ref{tab:case_study} illustrates such examples from our dataset\footnote{The table does not include paraphrases for questions and answers.}, where conversational context was incorporated to generate the verbalization. 
For instance, in the first conversational example, the user asks the question ``\textit{What was the birth city of Lionel Messi?}"; the dataset here includes the verbalized answer ``\textit{The birth city of Lionel Messi was Rosario, Santa Fe.}". Alongside the question and the verbalized answer, the dataset also provides paraphrases for both. In the next turn, we have the question ``\textit{Is \underline{he} a member of the Colombian National soccer team?}" where ``\textit{he}" refers to ``\textit{Lionel Messi}". For such examples, we either provide verbalized answers using the relevant pronoun (e.g ``\textit{No, \underline{he} is not a member of the Colombian National soccer team.}") or even the entity itself, (e.g. ``\textit{No, \underline{Lionel Messi} does not represent Colombia at the international level.}"). 
Similarly, on the next conversation, which belongs to the ``\textit{TV series}" domain, we provide such answers. More precisely, on turn four, the user asks the question ``\textit{And how many seasons did the show last?}" referring to the seed entity ``\textit{Dexter}" from first turn. Even in such scenarios, the dataset provides answers that also contain the entity (e.g. ``\textit{\underline{Dexter} is an 8 season television series.}").

\begin{table}[!t]
\small
\centering
\begin{tabular}{l|l}
\toprule
\textbf{Domain} & \textbf{Conversations} \\ \midrule
Soccer & \begin{tabular}[c]{@{}l@{}}Q1: What was the birth city of Lionel Messi?\\ A1: The birth city of Lionel Messi was\\ \MyIndent Rosario, Santa Fe.\\ Q2: Is he a member of the Colombian\\ \MyIndent National soccer team?\\ A2: No, Lionel Messi does not represent Colombia\\ \MyIndent at the international level.\\ Q3: Which national team is he a member of?\\ A3: He is an Argentina national team player.\\ Q4: Which of their goalkeepers is youngest? \\ A4: Juan Musso is the youngest goalkeeper.\\ Q5: When did he join the team?\\ A5: He joined the team in 2019.\end{tabular} \\ \midrule
TV series & \begin{tabular}[c]{@{}l@{}}Q1: What network was Dexter on?\\ A1: Dexter aired on Showtime.\\ Q2: What year did the show debut?\\ A2: The show debuted in 2007.\\ Q3: Who starred in it?\\ A3: Michael C. Hall was the show's star.\\ Q4: And how many seasons did the show last?\\ A4: Dexter is an 8 season television series.\\ Q5: What was the main location of it?\\ A5: The show was set in [Miami].\end{tabular} \\ \bottomrule
\end{tabular}
\caption{Verbal-ConvQuestions conversation examples. Each conversation in the dataset consists of five turns.}
\label{tab:case_study}
\vspace{-2em}
\end{table}

\section{Related Work}\label{sec:related_work}
Our work relates to previous approaches on answer verbalization for KGQA. However, we also briefly refer to existing approaches for answer generation on reading comprehension QA.
There exist three answer verbalization datasets for KGQA systems. The first dataset, named VQuAnDa~\cite{kacupaj2020vquanda}, extended an existing (complex) QA dataset by generating one verbalized response for each question. Following, the authors proposed~\cite{kacupaj2021paraqa} to add multiple paraphrased answer verbalizations and illustrated that multiple answers positively affect the performance of machine learning models. Work from~\citet{biswas2021vanilla} proposed answer verbalization on existing large-scale QA datasets only for simple questions. However, these datasets do not cover multiple question turns involving anaphora/ellipses. To the best of our knowledge, the here proposed Verbal-ConvQuestions is the first dataset that provides answer verbalization for conversational KGQA.

For reading comprehension approaches, work from ~\citet{baheti2020fluent} studied the task of generating fluent QA responses in the context of building conversational agents. The authors proposed a framework that modifies the SQuAD 2.0 dataset \cite{rajpurkar2018squad2} so that it includes conversational answers, which is used to train sequence-to-sequence based generation models. Another work~\cite{Peshterliev2021ConversationalAnswerGeneration} presents a BART-based~\cite{lewis2020bart} model for conversational answer generation and evaluate the validity of generated responses using NLI entailment. For question paraphrasing, we point readers to recent works by \cite{yu2020few,vakulenko2021question}.

\section{Conclusions and Predicted Impact}\label{sec:conclusions}
Conversational QA over KGs has been a trending research topic in scientific literature since \citet{saha2018csqa} introduced the first dataset in this domain. However, unlike task-oriented dialog systems \cite{liu2018end}, the lack of verbalized answers was a significant research gap in existing datasets, hindering the development of approaches involving more natural conversations. We introduce a dataset, Verbal-ConvQuestions, that contains verbalized answers and multiple paraphrased utterances for each conversation. We further provide experiments with several baseline models to generate the answer utterances. Our error analysis illustrates error rates for model mispredictions in different categories. We believe that our empirical study, which highlighted gaps in the state-of-the-art models for answer verbalization, will serve as a basis for researchers. At the same time, to develop novel techniques for the introduced task.



\begin{thebibliography}{34}


\ifx \showCODEN    \undefined \def \showCODEN     #1{\unskip}     \fi
\ifx \showDOI      \undefined \def \showDOI       #1{#1}\fi
\ifx \showISBNx    \undefined \def \showISBNx     #1{\unskip}     \fi
\ifx \showISBNxiii \undefined \def \showISBNxiii  #1{\unskip}     \fi
\ifx \showISSN     \undefined \def \showISSN      #1{\unskip}     \fi
\ifx \showLCCN     \undefined \def \showLCCN      #1{\unskip}     \fi
\ifx \shownote     \undefined \def \shownote      #1{#1}          \fi
\ifx \showarticletitle \undefined \def \showarticletitle #1{#1}   \fi
\ifx \showURL      \undefined \def \showURL       {\relax}        \fi
\providecommand\bibfield[2]{#2}
\providecommand\bibinfo[2]{#2}
\providecommand\natexlab[1]{#1}
\providecommand\showeprint[2][]{arXiv:#2}

\bibitem[\protect\citeauthoryear{Baheti, Ritter, and Small}{Baheti
  et~al\mbox{.}}{2020}]%
        {baheti2020fluent}
\bibfield{author}{\bibinfo{person}{Ashutosh Baheti}, \bibinfo{person}{Alan
  Ritter}, {and} \bibinfo{person}{Kevin Small}.}
  \bibinfo{year}{2020}\natexlab{}.
\newblock \showarticletitle{Fluent Response Generation for Conversational
  Question Answering}. In \bibinfo{booktitle}{\emph{Proceedings of the 58th
  Annual Meeting of the Association for Computational Linguistics}}.
  \bibinfo{publisher}{Association for Computational Linguistics},
  \bibinfo{address}{Online}, \bibinfo{pages}{191--207}.
\newblock
\urldef\tempurl%
\url{https://doi.org/10.18653/v1/2020.acl-main.19}
\showDOI{\tempurl}


\bibitem[\protect\citeauthoryear{Banerjee and Lavie}{Banerjee and
  Lavie}{2005}]%
        {banerjee2005meteor}
\bibfield{author}{\bibinfo{person}{Satanjeev Banerjee} {and}
  \bibinfo{person}{Alon Lavie}.} \bibinfo{year}{2005}\natexlab{}.
\newblock \showarticletitle{{METEOR}: An Automatic Metric for {MT} Evaluation
  with Improved Correlation with Human Judgments}. In
  \bibinfo{booktitle}{\emph{Proceedings of the {ACL} Workshop on Intrinsic and
  Extrinsic Evaluation Measures for Machine Translation and/or Summarization}}.
  \bibinfo{publisher}{Association for Computational Linguistics},
  \bibinfo{address}{Ann Arbor, Michigan}, \bibinfo{pages}{65--72}.
\newblock
\urldef\tempurl%
\url{https://www.aclweb.org/anthology/W05-0909}
\showURL{%
\tempurl}


\bibitem[\protect\citeauthoryear{Biswas, Dubey, Rony, and Lehmann}{Biswas
  et~al\mbox{.}}{2020}]%
        {biswas2021vanilla}
\bibfield{author}{\bibinfo{person}{Debanjali Biswas}, \bibinfo{person}{Mohnish
  Dubey}, \bibinfo{person}{Md~Rashad Al~Hasan Rony}, {and}
  \bibinfo{person}{Jens Lehmann}.} \bibinfo{year}{2020}\natexlab{}.
\newblock \showarticletitle{VANiLLa: Verbalised ANswers in Natural Language at
  Large scale}.
\newblock  (\bibinfo{year}{2020}).
\newblock
\urldef\tempurl%
\url{https://sda.tech/projects/vanilla/}
\showURL{%
\tempurl}


\bibitem[\protect\citeauthoryear{Bojar, Chatterjee, Federmann, Fishel, Graham,
  Haddow, Huck, Yepes, Koehn, Monz, et~al\mbox{.}}{Bojar et~al\mbox{.}}{2018}]%
        {bojar2018proceedings}
\bibfield{author}{\bibinfo{person}{Ond{\v{r}}ej Bojar}, \bibinfo{person}{Rajen
  Chatterjee}, \bibinfo{person}{Christian Federmann}, \bibinfo{person}{Mark
  Fishel}, \bibinfo{person}{Yvette Graham}, \bibinfo{person}{Barry Haddow},
  \bibinfo{person}{Matthias Huck}, \bibinfo{person}{Antonio~Jimeno Yepes},
  \bibinfo{person}{Philipp Koehn}, \bibinfo{person}{Christof Monz},
  {et~al\mbox{.}}} \bibinfo{year}{2018}\natexlab{}.
\newblock \showarticletitle{Proceedings of the Third Conference on Machine
  Translation: Shared Task Papers}. In \bibinfo{booktitle}{\emph{Proceedings of
  the Third Conference on Machine Translation: Shared Task Papers}}.
\newblock


\bibitem[\protect\citeauthoryear{Christmann, Saha~Roy, Abujabal, Singh, and
  Weikum}{Christmann et~al\mbox{.}}{2019}]%
        {christmann2019look}
\bibfield{author}{\bibinfo{person}{Philipp Christmann},
  \bibinfo{person}{Rishiraj Saha~Roy}, \bibinfo{person}{Abdalghani Abujabal},
  \bibinfo{person}{Jyotsna Singh}, {and} \bibinfo{person}{Gerhard Weikum}.}
  \bibinfo{year}{2019}\natexlab{}.
\newblock \showarticletitle{Look before You Hop: Conversational Question
  Answering over Knowledge Graphs Using Judicious Context Expansion}. In
  \bibinfo{booktitle}{\emph{Proceedings of the 28th ACM International
  Conference on Information and Knowledge Management}} (Beijing, China)
  \emph{(\bibinfo{series}{CIKM '19})}. \bibinfo{publisher}{Association for
  Computing Machinery}, \bibinfo{address}{New York, NY, USA},
  \bibinfo{pages}{729–738}.
\newblock
\showISBNx{9781450369763}
\urldef\tempurl%
\url{https://doi.org/10.1145/3357384.3358016}
\showDOI{\tempurl}


\bibitem[\protect\citeauthoryear{Devlin, Chang, Lee, and Toutanova}{Devlin
  et~al\mbox{.}}{2019}]%
        {devlin2019bert}
\bibfield{author}{\bibinfo{person}{Jacob Devlin}, \bibinfo{person}{Ming-Wei
  Chang}, \bibinfo{person}{Kenton Lee}, {and} \bibinfo{person}{Kristina
  Toutanova}.} \bibinfo{year}{2019}\natexlab{}.
\newblock \showarticletitle{{BERT}: Pre-training of Deep Bidirectional
  Transformers for Language Understanding}. In
  \bibinfo{booktitle}{\emph{Proceedings of the 2019 Conference of the North
  {A}merican Chapter of the Association for Computational Linguistics: Human
  Language Technologies, Volume 1 (Long and Short Papers)}}.
  \bibinfo{publisher}{Association for Computational Linguistics},
  \bibinfo{address}{Minneapolis, Minnesota}, \bibinfo{pages}{4171--4186}.
\newblock
\urldef\tempurl%
\url{https://doi.org/10.18653/v1/N19-1423}
\showDOI{\tempurl}


\bibitem[\protect\citeauthoryear{Edunov, Ott, Auli, and Grangier}{Edunov
  et~al\mbox{.}}{2018}]%
        {edunov2018understanding}
\bibfield{author}{\bibinfo{person}{Sergey Edunov}, \bibinfo{person}{Myle Ott},
  \bibinfo{person}{Michael Auli}, {and} \bibinfo{person}{David Grangier}.}
  \bibinfo{year}{2018}\natexlab{}.
\newblock \showarticletitle{Understanding Back-Translation at Scale}. In
  \bibinfo{booktitle}{\emph{Proceedings of the 2018 Conference on Empirical
  Methods in Natural Language Processing}}. \bibinfo{publisher}{Association for
  Computational Linguistics}, \bibinfo{address}{Brussels, Belgium},
  \bibinfo{pages}{489--500}.
\newblock
\urldef\tempurl%
\url{https://doi.org/10.18653/v1/D18-1045}
\showDOI{\tempurl}


\bibitem[\protect\citeauthoryear{Einolghozati, Gupta, Diedrick, and
  Gupta}{Einolghozati et~al\mbox{.}}{2020}]%
        {einolghozati2020sound}
\bibfield{author}{\bibinfo{person}{Arash Einolghozati}, \bibinfo{person}{Anchit
  Gupta}, \bibinfo{person}{Keith Diedrick}, {and} \bibinfo{person}{Sonal
  Gupta}.} \bibinfo{year}{2020}\natexlab{}.
\newblock \showarticletitle{Sound Natural: Content Rephrasing in Dialog
  Systems}. In \bibinfo{booktitle}{\emph{Proceedings of the 2020 Conference on
  Empirical Methods in Natural Language Processing (EMNLP)}}.
  \bibinfo{pages}{5101--5108}.
\newblock


\bibitem[\protect\citeauthoryear{Federmann, Elachqar, and Quirk}{Federmann
  et~al\mbox{.}}{2019}]%
        {federmann2019multilingual}
\bibfield{author}{\bibinfo{person}{Christian Federmann},
  \bibinfo{person}{Oussama Elachqar}, {and} \bibinfo{person}{Chris Quirk}.}
  \bibinfo{year}{2019}\natexlab{}.
\newblock \showarticletitle{Multilingual Whispers: Generating Paraphrases with
  Translation}. In \bibinfo{booktitle}{\emph{Proceedings of the 5th Workshop on
  Noisy User-generated Text (W-NUT 2019)}}. \bibinfo{publisher}{Association for
  Computational Linguistics}, \bibinfo{address}{Hong Kong, China},
  \bibinfo{pages}{17--26}.
\newblock
\urldef\tempurl%
\url{https://doi.org/10.18653/v1/D19-5503}
\showDOI{\tempurl}


\bibitem[\protect\citeauthoryear{{Fu}, {Qiu}, {Tang}, {Li}, {Yu}, and
  {Sun}}{{Fu} et~al\mbox{.}}{2020}]%
        {fu2020survey}
\bibfield{author}{\bibinfo{person}{Bin {Fu}}, \bibinfo{person}{Yunqi {Qiu}},
  \bibinfo{person}{Chengguang {Tang}}, \bibinfo{person}{Yang {Li}},
  \bibinfo{person}{Haiyang {Yu}}, {and} \bibinfo{person}{Jian {Sun}}.}
  \bibinfo{year}{2020}\natexlab{}.
\newblock \showarticletitle{{A Survey on Complex Question Answering over
  Knowledge Base: Recent Advances and Challenges}}.
\newblock \bibinfo{journal}{\emph{arXiv e-prints}}, Article
  \bibinfo{articleno}{arXiv:2007.13069} (\bibinfo{date}{July}
  \bibinfo{year}{2020}), \bibinfo{numpages}{arXiv:2007.13069}~pages.
\newblock
\showeprint[arxiv]{2007.13069}~[cs.CL]


\bibitem[\protect\citeauthoryear{Gehring, Auli, Grangier, Yarats, and
  Dauphin}{Gehring et~al\mbox{.}}{2017}]%
        {gehring2017conv}
\bibfield{author}{\bibinfo{person}{Jonas Gehring}, \bibinfo{person}{Michael
  Auli}, \bibinfo{person}{David Grangier}, \bibinfo{person}{Denis Yarats},
  {and} \bibinfo{person}{Yann~N. Dauphin}.} \bibinfo{year}{2017}\natexlab{}.
\newblock \showarticletitle{Convolutional Sequence to Sequence Learning}. In
  \bibinfo{booktitle}{\emph{Proceedings of the 34th International Conference on
  Machine Learning - Volume 70}} (Sydney, NSW, Australia)
  \emph{(\bibinfo{series}{ICML'17})}. \bibinfo{publisher}{JMLR.org},
  \bibinfo{pages}{1243–1252}.
\newblock


\bibitem[\protect\citeauthoryear{Kacupaj, Banerjee, Singh, and Lehmann}{Kacupaj
  et~al\mbox{.}}{2021a}]%
        {kacupaj2021paraqa}
\bibfield{author}{\bibinfo{person}{Endri Kacupaj}, \bibinfo{person}{Barshana
  Banerjee}, \bibinfo{person}{Kuldeep Singh}, {and} \bibinfo{person}{Jens
  Lehmann}.} \bibinfo{year}{2021}\natexlab{a}.
\newblock \showarticletitle{Para{\{}QA{\}}: A Question Answering Dataset with
  Paraphrase Responses for Single-Turn Conversation}. In
  \bibinfo{booktitle}{\emph{Eighteenth Extended Semantic Web Conference -
  Resources Track}}.
\newblock
\urldef\tempurl%
\url{https://openreview.net/forum?id=_GmNcCEAea3}
\showURL{%
\tempurl}


\bibitem[\protect\citeauthoryear{Kacupaj, Plepi, Singh, Thakkar, Lehmann, and
  Maleshkova}{Kacupaj et~al\mbox{.}}{2021b}]%
        {kacupaj2021lasagne}
\bibfield{author}{\bibinfo{person}{Endri Kacupaj}, \bibinfo{person}{Joan
  Plepi}, \bibinfo{person}{Kuldeep Singh}, \bibinfo{person}{Harsh Thakkar},
  \bibinfo{person}{Jens Lehmann}, {and} \bibinfo{person}{Maria Maleshkova}.}
  \bibinfo{year}{2021}\natexlab{b}.
\newblock \showarticletitle{Conversational Question Answering over Knowledge
  Graphs with Transformer and Graph Attention Networks}. In
  \bibinfo{booktitle}{\emph{Proceedings of the 16th Conference of the European
  Chapter of the Association for Computational Linguistics: Main Volume}}.
  \bibinfo{publisher}{Association for Computational Linguistics},
  \bibinfo{address}{Online}, \bibinfo{pages}{850--862}.
\newblock
\urldef\tempurl%
\url{https://www.aclweb.org/anthology/2021.eacl-main.72}
\showURL{%
\tempurl}


\bibitem[\protect\citeauthoryear{Kacupaj, Premnadh, Singh, Lehmann, and
  Maleshkova}{Kacupaj et~al\mbox{.}}{2021c}]%
        {kacupaj2021vogue}
\bibfield{author}{\bibinfo{person}{Endri Kacupaj}, \bibinfo{person}{Shyamnath
  Premnadh}, \bibinfo{person}{Kuldeep Singh}, \bibinfo{person}{Jens Lehmann},
  {and} \bibinfo{person}{Maria Maleshkova}.} \bibinfo{year}{2021}\natexlab{c}.
\newblock \showarticletitle{VOGUE: Answer Verbalization Through Multi-Task
  Learning}. In \bibinfo{booktitle}{\emph{Joint European Conference on Machine
  Learning and Knowledge Discovery in Databases}}. Springer,
  \bibinfo{pages}{563--579}.
\newblock


\bibitem[\protect\citeauthoryear{Kacupaj, Zafar, Lehmann, and
  Maleshkova}{Kacupaj et~al\mbox{.}}{2020}]%
        {kacupaj2020vquanda}
\bibfield{author}{\bibinfo{person}{Endri Kacupaj}, \bibinfo{person}{Hamid
  Zafar}, \bibinfo{person}{Jens Lehmann}, {and} \bibinfo{person}{Maria
  Maleshkova}.} \bibinfo{year}{2020}\natexlab{}.
\newblock \showarticletitle{VQuAnDa: Verbalization QUestion ANswering DAtaset}.
  In \bibinfo{booktitle}{\emph{The Semantic Web}},
  \bibfield{editor}{\bibinfo{person}{Andreas Harth}, \bibinfo{person}{Sabrina
  Kirrane}, \bibinfo{person}{Axel-Cyrille Ngonga~Ngomo}, \bibinfo{person}{Heiko
  Paulheim}, \bibinfo{person}{Anisa Rula}, \bibinfo{person}{Anna~Lisa Gentile},
  \bibinfo{person}{Peter Haase}, {and} \bibinfo{person}{Michael Cochez}}
  (Eds.). \bibinfo{publisher}{Springer International Publishing},
  \bibinfo{address}{Cham}, \bibinfo{pages}{531--547}.
\newblock
\showISBNx{978-3-030-49461-2}


\bibitem[\protect\citeauthoryear{Kaiser, Saha~Roy, and Weikum}{Kaiser
  et~al\mbox{.}}{2021}]%
        {conquer2021kaiser}
\bibfield{author}{\bibinfo{person}{Magdalena Kaiser}, \bibinfo{person}{Rishiraj
  Saha~Roy}, {and} \bibinfo{person}{Gerhard Weikum}.}
  \bibinfo{year}{2021}\natexlab{}.
\newblock \showarticletitle{Reinforcement Learning from Reformulations In
  Conversational Question Answering over Knowledge Graphs}. In
  \bibinfo{booktitle}{\emph{Proceedings of the 44th International ACM SIGIR
  Conference on Research and Development in Information Retrieval}}.
  \bibinfo{pages}{459–469}.
\newblock


\bibitem[\protect\citeauthoryear{Kingma and Ba}{Kingma and Ba}{2015}]%
        {kingma2015adam}
\bibfield{author}{\bibinfo{person}{Diederik~P Kingma} {and}
  \bibinfo{person}{Jimmy Ba}.} \bibinfo{year}{2015}\natexlab{}.
\newblock \showarticletitle{Adam, A method for stochastic optimization}. In
  \bibinfo{booktitle}{\emph{3rd ICLR}}.
\newblock


\bibitem[\protect\citeauthoryear{Lewis, Liu, Goyal, Ghazvininejad, Mohamed,
  Levy, Stoyanov, and Zettlemoyer}{Lewis et~al\mbox{.}}{2020}]%
        {lewis2020bart}
\bibfield{author}{\bibinfo{person}{Mike Lewis}, \bibinfo{person}{Yinhan Liu},
  \bibinfo{person}{Naman Goyal}, \bibinfo{person}{Marjan Ghazvininejad},
  \bibinfo{person}{Abdelrahman Mohamed}, \bibinfo{person}{Omer Levy},
  \bibinfo{person}{Veselin Stoyanov}, {and} \bibinfo{person}{Luke
  Zettlemoyer}.} \bibinfo{year}{2020}\natexlab{}.
\newblock \showarticletitle{{BART}: Denoising Sequence-to-Sequence Pre-training
  for Natural Language Generation, Translation, and Comprehension}. In
  \bibinfo{booktitle}{\emph{Proceedings of the 58th Annual Meeting of the
  Association for Computational Linguistics}}. \bibinfo{publisher}{Association
  for Computational Linguistics}, \bibinfo{address}{Online},
  \bibinfo{pages}{7871--7880}.
\newblock
\urldef\tempurl%
\url{https://doi.org/10.18653/v1/2020.acl-main.703}
\showDOI{\tempurl}


\bibitem[\protect\citeauthoryear{Liu and Lane}{Liu and Lane}{2018}]%
        {liu2018end}
\bibfield{author}{\bibinfo{person}{Bing Liu} {and} \bibinfo{person}{Ian Lane}.}
  \bibinfo{year}{2018}\natexlab{}.
\newblock \showarticletitle{End-to-end learning of task-oriented dialogs}. In
  \bibinfo{booktitle}{\emph{Proceedings of the 2018 Conference of the North
  American Chapter of the Association for Computational Linguistics: Student
  Research Workshop}}. \bibinfo{pages}{67--73}.
\newblock


\bibitem[\protect\citeauthoryear{Loshchilov and Hutter}{Loshchilov and
  Hutter}{2017}]%
        {loshchilov2017decoupled}
\bibfield{author}{\bibinfo{person}{Ilya Loshchilov} {and}
  \bibinfo{person}{Frank Hutter}.} \bibinfo{year}{2017}\natexlab{}.
\newblock \showarticletitle{Decoupled weight decay regularization}.
\newblock \bibinfo{journal}{\emph{arXiv preprint arXiv:1711.05101}}
  (\bibinfo{year}{2017}).
\newblock


\bibitem[\protect\citeauthoryear{Papineni, Roukos, Ward, and Zhu}{Papineni
  et~al\mbox{.}}{2002}]%
        {papineni2002bleu}
\bibfield{author}{\bibinfo{person}{Kishore Papineni}, \bibinfo{person}{Salim
  Roukos}, \bibinfo{person}{Todd Ward}, {and} \bibinfo{person}{Wei-Jing Zhu}.}
  \bibinfo{year}{2002}\natexlab{}.
\newblock \showarticletitle{{B}leu: a Method for Automatic Evaluation of
  Machine Translation}. In \bibinfo{booktitle}{\emph{Proceedings of the 40th
  Annual Meeting of the Association for Computational Linguistics}}.
  \bibinfo{publisher}{Association for Computational Linguistics},
  \bibinfo{address}{Philadelphia, Pennsylvania, USA},
  \bibinfo{pages}{311--318}.
\newblock
\urldef\tempurl%
\url{https://doi.org/10.3115/1073083.1073135}
\showDOI{\tempurl}


\bibitem[\protect\citeauthoryear{{Peshterliev}, {Oguz}, {Chatterjee}, {Inan},
  and {Bhardwaj}}{{Peshterliev} et~al\mbox{.}}{2021}]%
        {Peshterliev2021ConversationalAnswerGeneration}
\bibfield{author}{\bibinfo{person}{Stan {Peshterliev}}, \bibinfo{person}{Barlas
  {Oguz}}, \bibinfo{person}{Debojeet {Chatterjee}}, \bibinfo{person}{Hakan
  {Inan}}, {and} \bibinfo{person}{Vikas {Bhardwaj}}.}
  \bibinfo{year}{2021}\natexlab{}.
\newblock \showarticletitle{{Conversational Answer Generation and Factuality
  for Reading Comprehension Question-Answering}}.
\newblock \bibinfo{journal}{\emph{arXiv e-prints}}, Article
  \bibinfo{articleno}{arXiv:2103.06500} (\bibinfo{date}{March}
  \bibinfo{year}{2021}), \bibinfo{numpages}{arXiv:2103.06500}~pages.
\newblock
\showeprint[arxiv]{2103.06500}~[cs.CL]


\bibitem[\protect\citeauthoryear{Peskov, Clarke, Krone, Fodor, Zhang, Youssef,
  and Diab}{Peskov et~al\mbox{.}}{2019}]%
        {peskov2019multi}
\bibfield{author}{\bibinfo{person}{Denis Peskov}, \bibinfo{person}{Nancy
  Clarke}, \bibinfo{person}{Jason Krone}, \bibinfo{person}{Brigi Fodor},
  \bibinfo{person}{Yi Zhang}, \bibinfo{person}{Adel Youssef}, {and}
  \bibinfo{person}{Mona Diab}.} \bibinfo{year}{2019}\natexlab{}.
\newblock \showarticletitle{Multi-domain goal-oriented dialogues (multidogo):
  Strategies toward curating and annotating large scale dialogue data}. In
  \bibinfo{booktitle}{\emph{Proceedings of the 2019 Conference on Empirical
  Methods in Natural Language Processing and the 9th International Joint
  Conference on Natural Language Processing (EMNLP-IJCNLP)}}.
  \bibinfo{pages}{4518--4528}.
\newblock


\bibitem[\protect\citeauthoryear{Plepi, Kacupaj, Singh, Thakkar, and
  Lehmann}{Plepi et~al\mbox{.}}{2021}]%
        {plepi2021context}
\bibfield{author}{\bibinfo{person}{Joan Plepi}, \bibinfo{person}{Endri
  Kacupaj}, \bibinfo{person}{Kuldeep Singh}, \bibinfo{person}{Harsh Thakkar},
  {and} \bibinfo{person}{Jens Lehmann}.} \bibinfo{year}{2021}\natexlab{}.
\newblock \showarticletitle{Context transformer with stacked pointer networks
  for conversational question answering over knowledge graphs}. In
  \bibinfo{booktitle}{\emph{European Semantic Web Conference}}. Springer,
  \bibinfo{pages}{356--371}.
\newblock


\bibitem[\protect\citeauthoryear{Raffel, Shazeer, Roberts, Lee, Narang, Matena,
  Zhou, Li, and Liu}{Raffel et~al\mbox{.}}{2020}]%
        {colin2019t5}
\bibfield{author}{\bibinfo{person}{Colin Raffel}, \bibinfo{person}{Noam
  Shazeer}, \bibinfo{person}{Adam Roberts}, \bibinfo{person}{Katherine Lee},
  \bibinfo{person}{Sharan Narang}, \bibinfo{person}{Michael Matena},
  \bibinfo{person}{Yanqi Zhou}, \bibinfo{person}{Wei Li}, {and}
  \bibinfo{person}{Peter~J. Liu}.} \bibinfo{year}{2020}\natexlab{}.
\newblock \showarticletitle{Exploring the Limits of Transfer Learning with a
  Unified Text-to-Text Transformer}.
\newblock \bibinfo{journal}{\emph{Journal of Machine Learning Research}}
  \bibinfo{volume}{21}, \bibinfo{number}{140} (\bibinfo{year}{2020}),
  \bibinfo{pages}{1--67}.
\newblock
\urldef\tempurl%
\url{http://jmlr.org/papers/v21/20-074.html}
\showURL{%
\tempurl}


\bibitem[\protect\citeauthoryear{Rajpurkar, Jia, and Liang}{Rajpurkar
  et~al\mbox{.}}{2018}]%
        {rajpurkar2018squad2}
\bibfield{author}{\bibinfo{person}{Pranav Rajpurkar}, \bibinfo{person}{Robin
  Jia}, {and} \bibinfo{person}{Percy Liang}.} \bibinfo{year}{2018}\natexlab{}.
\newblock \showarticletitle{Know What You Don{'}t Know: Unanswerable Questions
  for {SQ}u{AD}}. In \bibinfo{booktitle}{\emph{Proceedings of the 56th Annual
  Meeting of the Association for Computational Linguistics (Volume 2: Short
  Papers)}}. \bibinfo{publisher}{Association for Computational Linguistics},
  \bibinfo{address}{Melbourne, Australia}, \bibinfo{pages}{784--789}.
\newblock
\urldef\tempurl%
\url{https://doi.org/10.18653/v1/P18-2124}
\showDOI{\tempurl}


\bibitem[\protect\citeauthoryear{Rothe, Narayan, and Severyn}{Rothe
  et~al\mbox{.}}{2020}]%
        {rothe2020bertseq2seq}
\bibfield{author}{\bibinfo{person}{Sascha Rothe}, \bibinfo{person}{Shashi
  Narayan}, {and} \bibinfo{person}{Aliaksei Severyn}.}
  \bibinfo{year}{2020}\natexlab{}.
\newblock \showarticletitle{Leveraging Pre-trained Checkpoints for Sequence
  Generation Tasks}.
\newblock \bibinfo{journal}{\emph{Transactions of the Association for
  Computational Linguistics}}  \bibinfo{volume}{8} (\bibinfo{year}{2020}),
  \bibinfo{pages}{264--280}.
\newblock
\urldef\tempurl%
\url{https://doi.org/10.1162/tacl_a_00313}
\showDOI{\tempurl}


\bibitem[\protect\citeauthoryear{Saha, Pahuja, Khapra, Sankaranarayanan, and
  Chandar}{Saha et~al\mbox{.}}{2018}]%
        {saha2018csqa}
\bibfield{author}{\bibinfo{person}{Amrita Saha}, \bibinfo{person}{Vardaan
  Pahuja}, \bibinfo{person}{Mitesh~M. Khapra}, \bibinfo{person}{Karthik
  Sankaranarayanan}, {and} \bibinfo{person}{Sarath Chandar}.}
  \bibinfo{year}{2018}\natexlab{}.
\newblock \showarticletitle{Complex Sequential Question Answering: Towards
  Learning to Converse Over Linked Question Answer Pairs with a Knowledge
  Graph}. In \bibinfo{booktitle}{\emph{Proceedings of the Thirty-Second {AAAI}
  Conference on Artificial Intelligence, (AAAI-18), the 30th innovative
  Applications of Artificial Intelligence (IAAI-18), and the 8th {AAAI}
  Symposium on Educational Advances in Artificial Intelligence (EAAI-18), New
  Orleans, Louisiana, USA, February 2-7, 2018}},
  \bibfield{editor}{\bibinfo{person}{Sheila~A. McIlraith} {and}
  \bibinfo{person}{Kilian~Q. Weinberger}} (Eds.). \bibinfo{publisher}{{AAAI}
  Press}, \bibinfo{pages}{705--713}.
\newblock
\urldef\tempurl%
\url{https://www.aaai.org/ocs/index.php/AAAI/AAAI18/paper/view/17181}
\showURL{%
\tempurl}


\bibitem[\protect\citeauthoryear{Singh, Radhakrishna, Both, Shekarpour, Lytra,
  Usbeck, Vyas, Khikmatullaev, Punjani, Lange, Vidal, Lehmann, and Auer}{Singh
  et~al\mbox{.}}{2018}]%
        {singh2018reinvent}
\bibfield{author}{\bibinfo{person}{Kuldeep Singh},
  \bibinfo{person}{Arun~Sethupat Radhakrishna}, \bibinfo{person}{Andreas Both},
  \bibinfo{person}{Saeedeh Shekarpour}, \bibinfo{person}{Ioanna Lytra},
  \bibinfo{person}{Ricardo Usbeck}, \bibinfo{person}{Akhilesh Vyas},
  \bibinfo{person}{Akmal Khikmatullaev}, \bibinfo{person}{Dharmen Punjani},
  \bibinfo{person}{Christoph Lange}, \bibinfo{person}{Maria~Esther Vidal},
  \bibinfo{person}{Jens Lehmann}, {and} \bibinfo{person}{S\"{o}ren Auer}.}
  \bibinfo{year}{2018}\natexlab{}.
\newblock \showarticletitle{Why Reinvent the Wheel: Let's Build Question
  Answering Systems Together}. In \bibinfo{booktitle}{\emph{Proceedings of the
  2018 World Wide Web Conference}} (Lyon, France) \emph{(\bibinfo{series}{WWW
  '18})}. \bibinfo{publisher}{International World Wide Web Conferences Steering
  Committee}, \bibinfo{address}{Republic and Canton of Geneva, CHE},
  \bibinfo{pages}{1247–1256}.
\newblock
\showISBNx{9781450356398}
\urldef\tempurl%
\url{https://doi.org/10.1145/3178876.3186023}
\showDOI{\tempurl}


\bibitem[\protect\citeauthoryear{Vakulenko, Longpre, Tu, and Anantha}{Vakulenko
  et~al\mbox{.}}{2021}]%
        {vakulenko2021question}
\bibfield{author}{\bibinfo{person}{Svitlana Vakulenko}, \bibinfo{person}{Shayne
  Longpre}, \bibinfo{person}{Zhucheng Tu}, {and} \bibinfo{person}{Raviteja
  Anantha}.} \bibinfo{year}{2021}\natexlab{}.
\newblock \showarticletitle{Question rewriting for conversational question
  answering}. In \bibinfo{booktitle}{\emph{Proceedings of the 14th ACM
  International Conference on Web Search and Data Mining}}.
  \bibinfo{pages}{355--363}.
\newblock


\bibitem[\protect\citeauthoryear{Vaswani, Shazeer, Parmar, Uszkoreit, Jones,
  Gomez, Kaiser, and Polosukhin}{Vaswani et~al\mbox{.}}{2017}]%
        {vaswani2017attention}
\bibfield{author}{\bibinfo{person}{Ashish Vaswani}, \bibinfo{person}{Noam
  Shazeer}, \bibinfo{person}{Niki Parmar}, \bibinfo{person}{Jakob Uszkoreit},
  \bibinfo{person}{Llion Jones}, \bibinfo{person}{Aidan~N. Gomez},
  \bibinfo{person}{undefinedukasz Kaiser}, {and} \bibinfo{person}{Illia
  Polosukhin}.} \bibinfo{year}{2017}\natexlab{}.
\newblock \showarticletitle{Attention is All You Need}. In
  \bibinfo{booktitle}{\emph{Proceedings of the 31st International Conference on
  Neural Information Processing Systems}} (Long Beach, California, USA)
  \emph{(\bibinfo{series}{NIPS'17})}. \bibinfo{publisher}{Curran Associates
  Inc.}, \bibinfo{address}{Red Hook, NY, USA}, \bibinfo{pages}{6000–6010}.
\newblock
\showISBNx{9781510860964}


\bibitem[\protect\citeauthoryear{Vrande\v{c}i\'{c} and
  Kr\"{o}tzsch}{Vrande\v{c}i\'{c} and Kr\"{o}tzsch}{2014}]%
        {vrande2014wikidata}
\bibfield{author}{\bibinfo{person}{Denny Vrande\v{c}i\'{c}} {and}
  \bibinfo{person}{Markus Kr\"{o}tzsch}.} \bibinfo{year}{2014}\natexlab{}.
\newblock \showarticletitle{Wikidata: A Free Collaborative Knowledgebase}.
\newblock \bibinfo{journal}{\emph{Commun. ACM}} \bibinfo{volume}{57},
  \bibinfo{number}{10} (\bibinfo{date}{Sept.} \bibinfo{year}{2014}),
  \bibinfo{pages}{78–85}.
\newblock
\showISSN{0001-0782}
\urldef\tempurl%
\url{https://doi.org/10.1145/2629489}
\showDOI{\tempurl}


\bibitem[\protect\citeauthoryear{Wolf, Debut, Sanh, Chaumond, Delangue, Moi,
  Cistac, Rault, Louf, Funtowicz, Davison, Shleifer, von Platen, Ma, Jernite,
  Plu, Xu, Scao, Gugger, Drame, Lhoest, and Rush}{Wolf et~al\mbox{.}}{2020}]%
        {wolf2020transformers}
\bibfield{author}{\bibinfo{person}{Thomas Wolf}, \bibinfo{person}{Lysandre
  Debut}, \bibinfo{person}{Victor Sanh}, \bibinfo{person}{Julien Chaumond},
  \bibinfo{person}{Clement Delangue}, \bibinfo{person}{Anthony Moi},
  \bibinfo{person}{Pierric Cistac}, \bibinfo{person}{Tim Rault},
  \bibinfo{person}{Rémi Louf}, \bibinfo{person}{Morgan Funtowicz},
  \bibinfo{person}{Joe Davison}, \bibinfo{person}{Sam Shleifer},
  \bibinfo{person}{Patrick von Platen}, \bibinfo{person}{Clara Ma},
  \bibinfo{person}{Yacine Jernite}, \bibinfo{person}{Julien Plu},
  \bibinfo{person}{Canwen Xu}, \bibinfo{person}{Teven~Le Scao},
  \bibinfo{person}{Sylvain Gugger}, \bibinfo{person}{Mariama Drame},
  \bibinfo{person}{Quentin Lhoest}, {and} \bibinfo{person}{Alexander~M. Rush}.}
  \bibinfo{year}{2020}\natexlab{}.
\newblock \showarticletitle{Transformers: State-of-the-Art Natural Language
  Processing}. In \bibinfo{booktitle}{\emph{EMNLP 2020: System
  Demonstrations}}. \bibinfo{publisher}{Association for Computational
  Linguistics}, \bibinfo{address}{Online}, \bibinfo{pages}{38--45}.
\newblock


\bibitem[\protect\citeauthoryear{Yu, Liu, Yang, Xiong, Bennett, Gao, and
  Liu}{Yu et~al\mbox{.}}{2020}]%
        {yu2020few}
\bibfield{author}{\bibinfo{person}{Shi Yu}, \bibinfo{person}{Jiahua Liu},
  \bibinfo{person}{Jingqin Yang}, \bibinfo{person}{Chenyan Xiong},
  \bibinfo{person}{Paul Bennett}, \bibinfo{person}{Jianfeng Gao}, {and}
  \bibinfo{person}{Zhiyuan Liu}.} \bibinfo{year}{2020}\natexlab{}.
\newblock \showarticletitle{Few-Shot Generative Conversational Query
  Rewriting}. In \bibinfo{booktitle}{\emph{Proceedings of the 43rd
  International ACM SIGIR Conference on Research and Development in Information
  Retrieval}}. \bibinfo{pages}{1933--1936}.
\newblock


\end{thebibliography}
\end{document}